\documentclass[runningheads]{llncs}

\usepackage{eccv}

\usepackage{eccvabbrv}

\usepackage{graphicx}
\usepackage{booktabs}

\usepackage[accsupp]{axessibility}  

\usepackage[pagebackref,breaklinks,colorlinks,citecolor=eccvblue]{hyperref}

\usepackage{orcidlink}

\usepackage{overpic}
\usepackage{multirow}
\usepackage{wrapfig}
\usepackage{enumitem}

\begin{document}

\title{GUICrafter: Weakly-Supervised GUI Agent Leveraging Massive Unannotated Screenshots} 

\titlerunning{GUICrafter}

\author{Sunqi Fan\inst{1} \and
Lingshan Chen\inst{1} \and
Runqi Yin\inst{1} \and
Qingle Liu\inst{1} \and
Yongming Rao\inst{2} \and
Meng-Hao Guo\inst{1} \and
Shi-Min Hu\inst{1}}

\authorrunning{S. Fan et al.}

\institute{Tsinghua University\and
Tencent Hunyuan\\
\email{\{fansq20, cls24, yrq24, lql24\}@mails.tsinghua.edu.cn, 
\\raoyongming95@gmail.com,
\\\{gmh, shimin\}@tsinghua.edu.cn}\\
}

\maketitle

\begin{abstract}
Data, as the fundamental substrate of modern intelligence, has greatly driven the development of current foundation models. 
Naturally, researchers aim to extend this paradigm to the domain of GUI agents, hoping to build strong GUI agents through a similar paradigm.
However, GUI agent data cannot be directly harvested from the internet, making it costly and difficult to collect at scale.
As a result, current GUI agents suffer from poor cross-device generalization and limited visual grounding ability for fine-grained GUI elements.
As an attempt to address data challenge in GUI agents, we propose GUICrafter,
a weakly-supervised GUI agent leveraging massive unannotated screenshots to substantially reduce the reliance on expensive human annotations. 
GUICrafter explores a curriculum learning framework for training GUI agents through two progressive stages.
First, the model learns visual grounding from large-scale unannotated screenshots and webpages, leveraging the rich contextual signals inherent in GUI interactions without human annotations.
Then, in Stage 2, we leverage a small amount of high-quality data to calibrate the model via reinforcement learning.
Experiments show that GUICrafter achieves competitive, or even superior, performance to advanced systems like UI-TARS while using only 0.1\% of its data.
Furthermore, under the same amount of annotated data, GUICrafter surpasses all previous methods such as GUI-R1.
Code, data, and models are available at \url{https://github.com/fansunqi/GUICrafter}.
  \keywords{GUI Agent \and Reinforcement Learning \and Weakly-Supervise}
\end{abstract}

\section{Introduction}
\label{sec:intro}

\begin{quote}
    \textit{“Nothing has such power to broaden the mind as the ability to investigate systematically and truly all that comes under thy observation in life.”}
    \begin{flushright}
        \textit{–- Marcus Aurelius}
    \end{flushright}
\end{quote}

\noindent GUI agents are powered by foundation models and can autonomously perform GUI interactions, simulating human operations such as clicking, typing, dragging, etc, to accomplish user-specified tasks on electronic devices.
Currently, most approaches rely on large-scale GUI task data to fine-tune Multimodal Large Language Models (MLLMs). However, annotating fine-grained grounding positions and multi-step actions for GUI tasks is highly labor-intensive, and existing MLLM-based automatic annotation methods often produce unreliable or low-quality labels, making it difficult to collect data at scale.


\begin{figure}[t]
  \centering
  
  \begin{minipage}{0.68\linewidth}
    \centering
    \includegraphics[width=\linewidth]{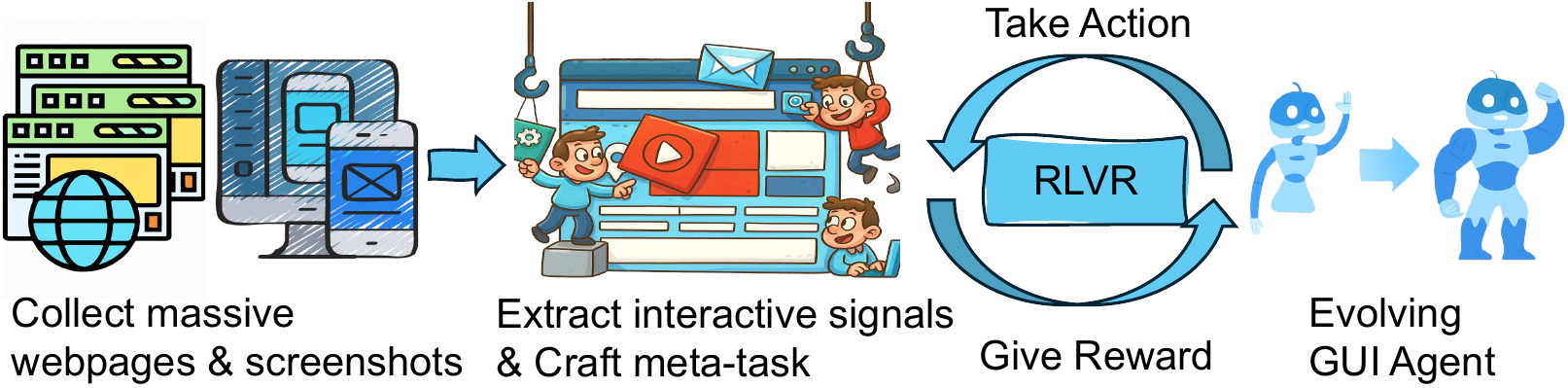}
  \end{minipage}
  \hfill
  \begin{minipage}{0.28\linewidth}
    \centering
    \footnotesize
    \begin{overpic}[width=\linewidth]{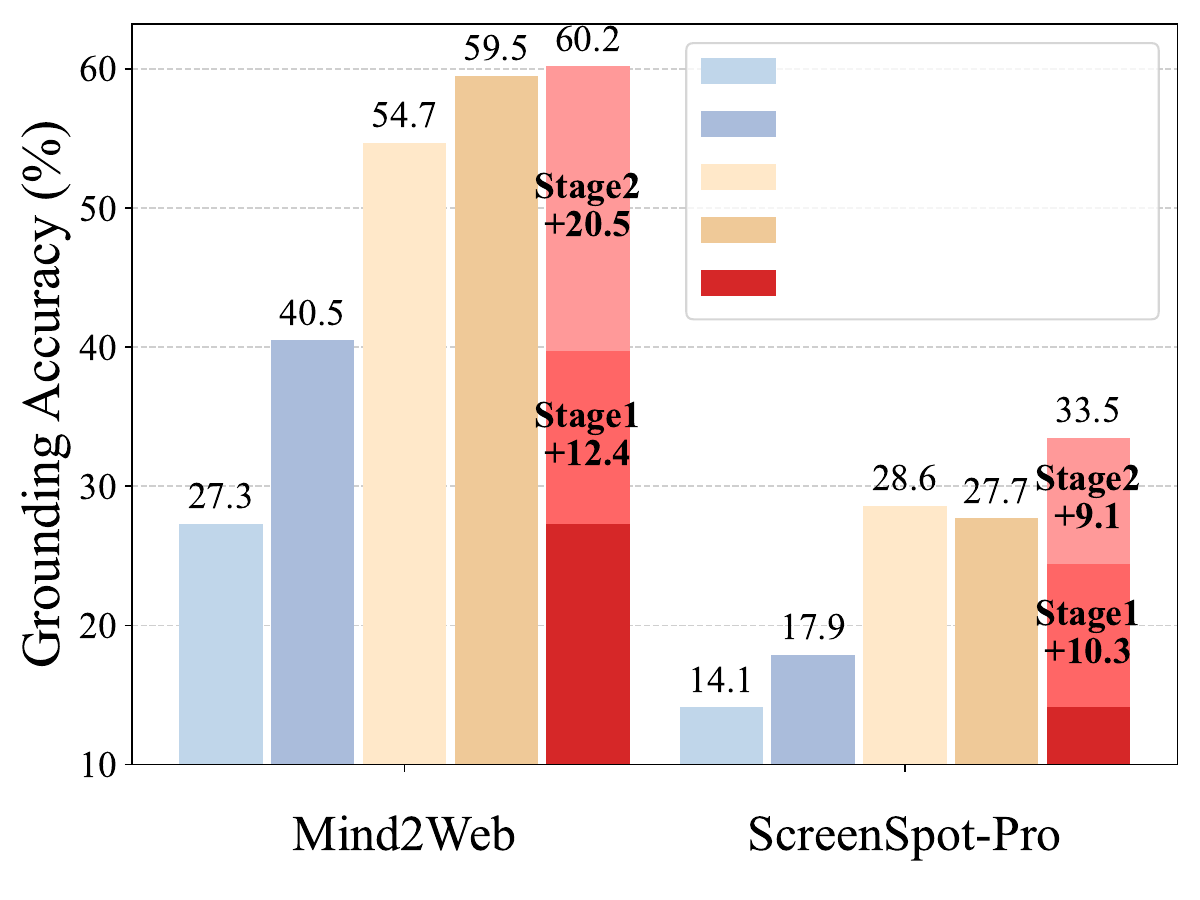}
      \put(66, 68){\scalebox{0.35}{Qwen2.5-VL-3B~\cite{bai2025qwen25vltechnicalreport}}}
      \put(66, 63.5){\scalebox{0.35}{UI-R1-3B~\cite{lu2025uir1enhancingefficientaction}}}
      \put(66, 59){\scalebox{0.35}{GUI-R1-3B~\cite{luo2025gui}}}
      \put(66, 54.5){\scalebox{0.35}{UI-TARS-2B~\cite{qin2025uitarspioneeringautomatedgui}}}
      \put(66, 50){\scalebox{0.35}{GUICrafter-3B (ours)}}
    \end{overpic}
  \end{minipage}
  
  \caption{Left: The pipeline of our Stage 1 weakly-supervised GUI pretraining, including data preparation and training process. 
  Right: Our GUICrafter model achieves a higher average grounding accuracy than all baselines on both Mind2Web~\cite{deng2023mind2web} and ScreenSpot-Pro~\cite{li2025screenspotproguigroundingprofessional} benchmarks. The results of GUI-R1 is reproduced using the same amount of annotated training data. We also highlight the significant improvements brought by Stage 1 and Stage 2 respectively.}
  \label{fig:motivation_combined}
\end{figure}
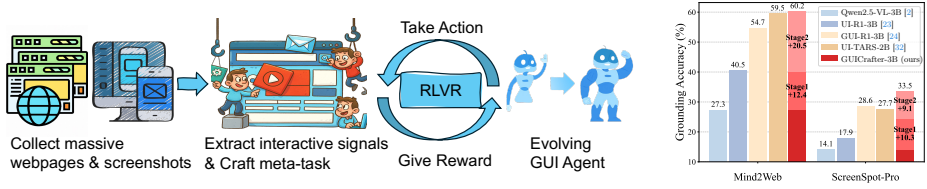

Due to insufficient and incomprehensive training data, current GUI agents face the following two main predicaments: \textbf{(1) GUI agents still struggle with visual grounding, as they may overlook subtle details of the screenshot.} 
The inability to accurately locate GUI elements often leads to failures in almost all GUI tasks.
\textbf{(2) GUI agents' cross-GUI and cross-domain generalization capability is limited.} They only perform well on the types of GUI covered in the training data. The root cause of both the visual grounding and generalization issues lies in the lack of comprehensive and diverse GUI-related training data, which fails to encompass the wide variety of GUI styles and design patterns. Without more varied and extensive data,
GUI agents are unable to fully understand and adapt to the vast range of GUI interface layouts they may encounter in real-world applications.

Confronted with the aforementioned challenges, we consider the following questions: 
can we leverage the vast amount of unannotated screenshot data from webpages and electronic devices to enhance the visual grounding ability and generalization of GUI agents? Furthermore, 
can we utilize the interaction signals within webpages or electronic device GUIs to provide appropriate feedback to GUI agents, thereby facilitating their improvement?



Building on this motivation, we propose the GUICrafter, enabling GUI agents to generalize their capabilities by observing massive amounts of unannotated screenshots. GUICrafter transforms unannotated screenshots into trainable data, offering an efficient and cost-effective data generation approach, as shown in Figure \ref{fig:motivation_combined}. It first collects massive webpages and uses web scraping tools to automatically extract interactive signals from these pages. It also leverages abundant mobile device pages from existing open-source datasets, which are richly annotated with automatically collected interactive GUI elements. It futher crafts corresponding meta-tasks for these interactive signals, which serve as an inductive summary of all GUI tasks. In contrast to traditional manually annotated tasks and ground-truth actions, GUICrafter replaces them with meta-tasks and corresponding actions derived from interactive signals, thus eliminating the need for labor-intensive annotation.

With the automatically generated data, we adopt a curriculum learning framework consisting of two progressive stages, using the Reinforcement Learning with Verifiable Rewards (RLVR) algorithm. In Stage 1, when the agent is presented with a meta-task, it should locate the corresponding interactive elements on the webpage, perform the appropriate action, and receive a reward, which in turn updates the parameters of the model, enabling it to evolve. 
In Stage 2, we incorporate high-quality manually annotated GUI task data to further enhance our agent model.

By using a curriculum learning strategy, better performance can be achieved. We have validated the effectiveness of both Stage 1 and the whole two-stage training process across six benchmarks of different platforms. Our evaluation results show that GUICrafter enhances the model's visual grounding and generalization capability significantly. As illustrated in Figure \ref{fig:motivation_combined}, both Stage 1 and Stage 2 lead to significant improvements in the grounding accuracy on Mind2Web~\cite{deng2023mind2web} and ScreenSpot-Pro~\cite{li2025screenspotproguigroundingprofessional} benchmarks. Additionally, GUICrafter exhibits high data efficiency and scalability. The total number of training data used in our Stage 1 and Stage 2 is only approximately 0.1\% of UI-TARS~\cite{qin2025uitarspioneeringautomatedgui}, yet we have achieved comparable or even superior performance. Furthermore, under the same amount of human-annotated data, GUICrafter surpasses all previous RLVR methods such as GUI-R1~\cite{luo2025gui}.  
Contributions can be summarized as follows:

\begin{itemize}
    \item We conduct an in-depth analysis of the data dilemma currently faced by GUI agents and propose a solution that leverages large amounts of unlabeled data together with a small amount of high-quality data. Correspondingly, we construct datasets following this design.
    \item Building upon the proposed data, we develop a two-stage reinforcement learning framework that 
    integrates weakly-supervised and supervised training to enhance data utilization and  learning efficiency.
    \item Experimental results demonstrate that, with our proposed data and algorithm, GUICrafter achieves superior performance compared to advanced GUI agent systems such as UI-TARS and GUI-R1. All the data, code, and models will be publicly released.
\end{itemize}

\section{Related Work}

\subsection{GUI Agents}

The development of Computer Use Agents (CUA) has long relied on using textual representations like HTML and accessibility trees~\cite{DBLP:conf/iclr/LiuGPSL18,deng2023mind2web,zhou2023webarena}. However, the text-based methods has been proven to have issues such as inconsistency, volatility, and limited scalability~\cite{xu2024aguvis}. These challenges have led to the rise of vision-based GUI agents that take screenshots as their input instead of textual descriptions~\cite{Hong_2024_CVPR, gou2024uground, gu2025uivenustechnicalreportbuilding, Yang_2025_CVPR, cheng-etal-2024-seeclick, wu2024atlas, qian-etal-2024-visual, wu2025guiactorcoordinatefreevisualgrounding, yuan2025enhancingvisualgroundinggui}. 
For example, Show-UI~\cite{Lin2025showui} uses UI-guided token selection to cut computational costs while achieving high zero-shot accuracy. UI-TARS~\cite{qin2025uitarspioneeringautomatedgui} achieves leading performance by integrating enhanced perception, unified action modeling, and system-2 reasoning. Many GUI agent datasets and benchmarks have also emerged~\cite{chen2024gui,davydova2025osuniverse,pan2024webcanvas,pandit2025synthesizing,deng2023mind2web,li2025screenspotproguigroundingprofessional,li2024effectsdatascaleui,kapoor2024omniact}. Industry has also introduced numerous GUI agent products that primarily rely on massive amounts of high-quality data, well-established sandbox infrastructures, and comprehensive curriculum training pipelines, such as UI-TARS-2~\cite{wang2025ui}, UI-Venus~\cite{gu2025uivenustechnicalreportbuilding}, MAI-UI~\cite{zhou2025maiuitechnicalreportrealworld}, Mobile-Agent~\cite{ye2025mobileagentv3fundamentalagentsgui}, AutoGLM~\cite{liu2024autoglm} and others~\cite{zeng2025uitron, lai2025computerrl, xu2026mobilerl}.

\subsection{Reinforcement Learning in GUI Agents}

Reinforcement Learning (RL) studies how agents learn through trial and error guided by reward signals.
RL has been extensively applied to GUI agents. Early work used workflow-guided exploration to collect valid web interaction trajectories~\cite{DBLP:conf/iclr/LiuGPSL18}. 
Recent studies advance along two lines: 

(1) Reinforcement-based visual grounding and refined reward modeling to improve data efficiency~\cite{yuan2025enhancingvisualgroundinggui, tang2025gui, lu2025orcust}, such as GUI-R1~\cite{luo2025gui} and UI-R1~\cite{lu2025uir1enhancingefficientaction}. \textbf{The key distinction between this work and prior efforts such as GUI-R1 and UI-R1 is that we introduce a weakly-supervised, annotation-free GUI pretraining stage.} This Stage 1 effectively enhances the GUI agent’s knowledge and comprehension of GUIs. 

(2) Online, multi-turn and self-evolving RL frameworks leveraging reward model and interactive environments, such as ZeroGUI~\cite{yang2025zeroguiautomatingonlinegui}, WebEvolver~\cite{fang2025webevolver}, MobileGUI-RL~\cite{DBLP:journals/corr/abs-2507-05720}, SEAgent~\cite{sun2025seagentselfevolvingcomputeruse} and WebAgent-R1~\cite{wei-etal-2025-webagent}. 
These methods typically depend on two forms of external information: 
(\romannumeral 1) task instruction annotation and (\romannumeral 2)  carefully constructed reward models, world models or handcrafted reward rubrics. 
Such designs often introduce additional human-annotation or LLM-generation cost, environment-specific assumptions, and limited scalability. As shown in Table \ref{tab:gui_rl_comparison}, \textbf{our method differs fundamentally in that (\romannumeral 1) it leverages simple meta-tasks instead of annotated tasks and (\romannumeral 2) it leverages the interaction of web and mobile platforms as reward signals instead of powerful reward models, world models or handcrafted reward rubrics.} This makes our framework conceptually simpler, easier to scale, and more transferable to real-world GUI environments.

\begin{table}[h]
\centering
\caption{\textbf{Comparison between our method and prior GUI-agent RL approaches.}
Prior methods typically rely on explicit task annotation and additional reward or world modeling, while our method uses meta-task and avoids explicit reward or world model design.
}
\label{tab:gui_rl_comparison}
\resizebox{\linewidth}{!}{%
\begin{tabular}{lcc}
\toprule
\textbf{Method} & \textbf{Task Source} & \textbf{Reward Signal} \\
\midrule
SE Agent~\cite{sun2025seagentselfevolvingcomputeruse}      & LLM-generated           & Reward Model \\
MobileGUI-RL~\cite{xu2026mobilerl}  & LLM-generated           & World State Model \\
ZeroGUI~\cite{yang2025zeroguiautomatingonlinegui}       & LLM-generated           & Reward Model \\
WebAgent-R1~\cite{wei-etal-2025-webagent}   & Human-annotated         & Rule-based Reward \\
\midrule
\textbf{GUICrafter} (ours) & \textbf{Meta-Task} & \textbf{Environment Interaction Signal} \\
\bottomrule
\end{tabular}%
}
\end{table}





\section{Method}
We present the preliminaries of our method (the GUI agent formulation and the GRPO algorithm) in \textbf{Appendix A}. Below, we first describe Stage 1 Weakly-Supervised GUI Pretraining, followed by Stage 2 High-Quality Reinforcement Fine-tuning.

\subsection{Weakly-Supervised GUI Pretraining}

\begin{figure*}[t]
  \centering
   \includegraphics[width=1.0\linewidth]{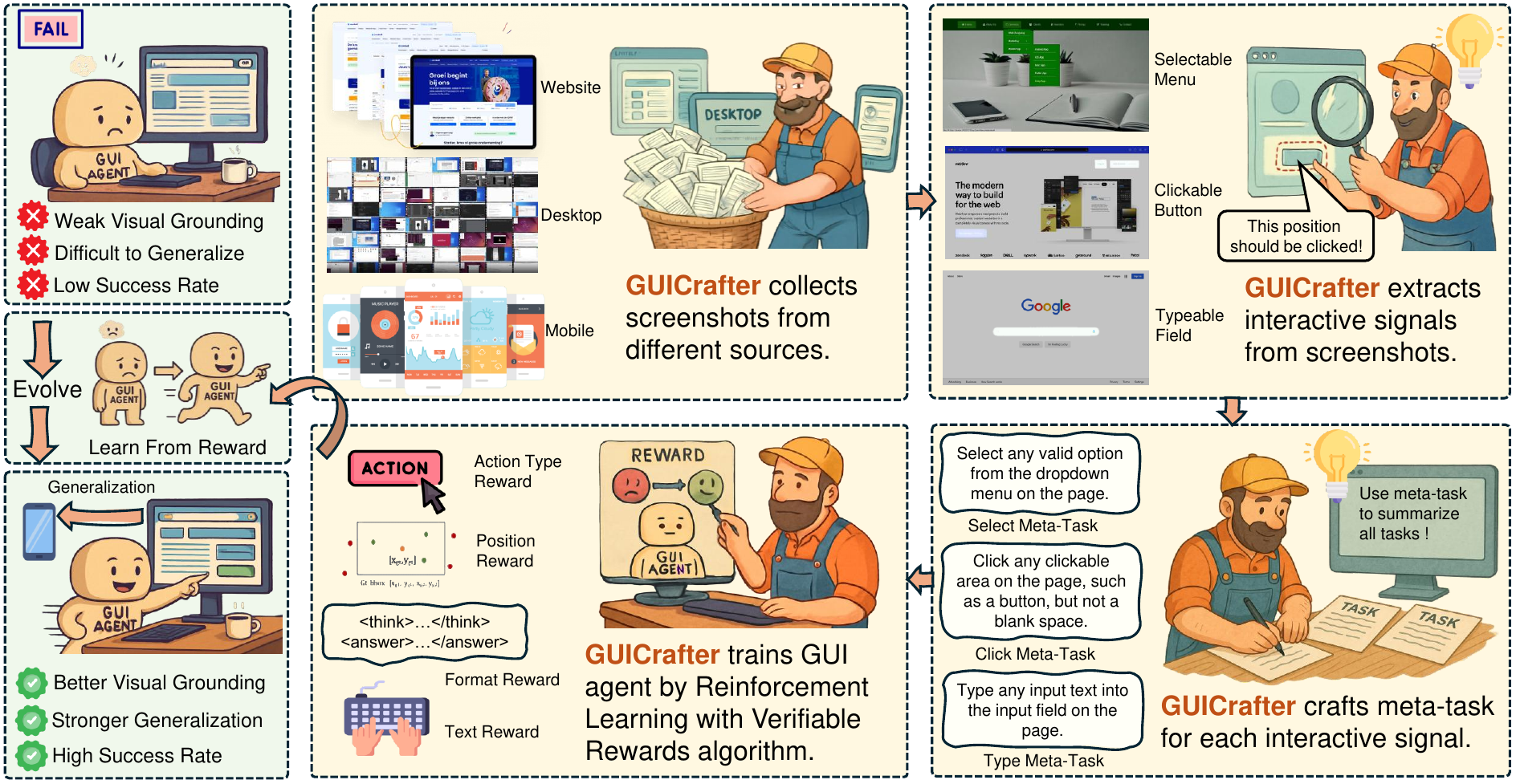}

   \caption{In Stage 1, we first collect GUI screenshots, extract interactive signals and craft meta-tasks. Meta-tasks can be viewed as an abstraction of human-annotated GUI tasks. The figure shows the interactive signals and meta-tasks for the website platform. Then, we use RLVR algorithm to train the GUI agent. This stage successfully enhances the agent’s visual grounding and generalization ability.}
   \label{fig:main}
\end{figure*}

The motivation of Stage 1 is to enable the GUI agent model to see and learn from a large number of GUI screenshots, similar to how LLMs read and learn from vast amounts of text during the pretraining phase. Therefore, we refer to this stage as “weakly-supervised GUI pretraining”. The overall pipeline of Stage 1 is illustrated in Figure \ref{fig:main}.

\paragraph{\textbf{Collecting Real-World Webpages \& Screenshots}} Our key insight is to leverage the implicit interaction signals embedded in real-world webpages and screenshots to enable the GUI agent to evolve without relying on extensive human annotations. Therefore, we first collect a large amount of real-world webpages and screenshots. 

For the web platform, we crawl abundant webpages of popular websites. To prevent the loss of image resources, we store the webpages in MHTML\footnote{MHTML is a web archiving file format used to combine, in a single computer file, the HTML code and its companion resources.} format. Since webpages within a complete multi-step GUI task are often interrelated, we adopt a similar strategy during crawling: starting from a portal site, detecting all URL links on the page, visiting each corresponding webpage recursively. This process naturally forms a tree-structured search pattern that simulates the distribution of webpages in real GUI tasks. In addition, we apply several autonomous filtering rules to the collected data, such as prioritizing English-language websites and filtering out those with pop-up windows.

For the mobile device platform, we utilize two large-scale open-source datasets AndroidControl~\cite{li2024effectsdatascaleui} and AITZ~\cite{zhang2024android} for their abundant page screenshots and wide span across various apps. Without using their human-annotated GUI task trajectories, we only utilize the automatically collected interactive GUI elements. 
We selected screenshots from each dataset and mixed them together to create the weakly-supervised mobile training data. 

\paragraph{\textbf{Crafting Interactive Signals \& Meta-Tasks}} After collecting abundant webpages and screenshots, we need to extract the embedded interactive signals, which can provide feedback to the GUI agent and facilitate its evolution.

For the web platform, we categorize the interactions between humans or agents and webpages into three actions: click, type, and select. Correspondingly, we employ browser simulation packages such as Playwright\footnote{Playwright is an open-source automation library for browser testing and web scraping developed by Microsoft.} in Python to identify clickable, typable, and selectable GUI elements from the MHTML files, and store them in the form of bounding box positions. We further craft three corresponding meta-tasks based on the three types of actions. For example, for the click action, the designed meta-task is \textit{“click any clickable area on the page, such as a button, but not a blank space”}. 
The meta-task abstracts all tasks related to a specific action and serves as the counterpart of manually annotated tasks. When the GUI agent receives a meta-task, it is expected to perform the corresponding action, with its output coordinates falling within the corresponding interactive regions, e.g., the clickable button.

For the mobile device platform, we extract the clickable, checkable and editable elements from each page's accessibility tree, and apply several automatic rules to filter out misannotated pages. 
We similarly craft corresponding meta-tasks for different types of interactive GUI elements. For the screenshots from AndroidControl~\cite{li2024effectsdatascaleui} dataset, the meta-task is to click on any clickable, checkable or editable element on the page. For the screenshots from AITZ~\cite{zhang2024android} dataset, the meta-task is to click on any icon element.

\paragraph{\textbf{Training Algorithm and Reward Design}}

\label{sec:stage1}

We integrate the feedback derived from these interaction signals into the Reinforcement Learning with Verifiable Rewards (RLVR) paradigm, enabling the GUI agent to evolve through efficient reinforcement learning algorithms.

We train the agent model to reason similar to OpenAI o1~\cite{o1blog} and DeepSeek-R1~\cite{guo2025deepseek}. The model first generates a reasoning process enclosed between two thinking tags, followed by the final answer enclosed between two answer tags. The final answer must be formatted in JSON, containing three key fields: action type, predicted position, and optional input text. If the model’s output strictly adheres to this format, it receives a format reward $R_{f}$ of 1; otherwise, the format reward $R_{f}$ is 0.

 For action modeling, we define a unified action space across different datasets and benchmarks. An action type reward $R_{type}$ of 1 is assigned only when the predicted action matches the ground-truth action; otherwise, the action type reward $R_{type}$ is set to 0.

For position prediction, a straightforward approach is to assign a position reward $R_p$ of 1 if the predicted point falls within the corresponding interactive region. For example, in the case of a click meta-task, when the predicted point lies within the bounding box of a clickable GUI element, the position reward $R_p$ is 1.
However, the interactive regions typically occupy a relatively large portion of the screenshot, which may lead to overly lenient positive feedback conditions and an imbalance between positive and negative rewards. To address this issue, we instead define the position reward $R_p$ based on the distance between the predicted point and the Gaussian distribution derived from the nearest interactive bounding box, similar to GUI-G$^2$~\cite{tang2025gui}. Specifically, given a predicted point $\mu_p=(c^p_x, c^p_y)$ and ground truth center $\mu_{gt} = (c^{gt}_x, c^{gt}_y)$, we compute the position reward $R_p$ using Equation \ref{eq:reward_point}:

\begin{equation}
\label{eq:reward_point}
R_{p} = \mathcal{N}(\mu_p; \mu_{gt}, \Sigma_{gt})
= \exp \left(
-\frac{1}{2}
\left[
\frac{(c_x^{p} - c_x^{gt})^2}{(\sigma_x^{gt})^2}
+
\frac{(c_y^{p} - c_y^{gt})^2}{(\sigma_y^{gt})^2}
\right]
\right)
\end{equation}

Here, $\Sigma_{gt} =
\begin{pmatrix}
(\sigma_x^{gt})^2 & 0 \\
0 & (\sigma_y^{gt})^2
\end{pmatrix}$ is the diagonal covariance matrix, where $\sigma_x^{gt}$ and $\sigma_y^{gt}$ are obtained by scaling the width and height of the bounding box with a hyperparameter factor, respectively.

For type and select actions that require input text, a text reward $R_{text}$ of 1 is assigned only when the predicted text is sufficiently similar to the ground-truth text, i.e.,  the token-level F1 score exceeds a predefined threshold; otherwise, the text reward $R_{text}$ is set to 0. 

Finally, the action reward $R_a$ is defined by Equation \ref{eq:reward_action} and the overall reward $R$ is computed using Equation \ref{eq:reward_overall}, where $\beta$ is a hyperparameter that balances the action reward and the format reward.

\begin{equation}
\label{eq:reward_action}
R_a =
\begin{cases}
R_{type} R_p, & \text{if task does not require text input}, \\[4pt]
R_{type} R_p R_{text}, & \text{if task requires text input}.
\end{cases}
\end{equation}

\begin{equation}
  R = R_a + \beta R_f
  \label{eq:reward_overall}
\end{equation}

After computing rewards, we optimize the base language model using GRPO algorithm described in \textbf{Appendix A}.

\subsection{High-Quality Reinforcement Fine-tuning}

In Stage 2, we collected a set of high-quality data and refined it through an LLM-assisted and rule-based filtering process with human verification, resulting in a cleaner and more reliable dataset. For both the web and desktop platform, we first applied the following rules to process the Mind2Web~\cite{deng2023mind2web} training dataset:

\begin{itemize}
    \item For cases where the rendered webpage does not match the original screenshot, we re-capture screenshots or filter out instances where the rendered webpage is incomplete or obscured by pop-up windows.
    \item We filter out samples with unclear task descriptions.
    \item We also filter out data whose action history included the above-mentioned invalid entries.
    \item Some bounding box annotations were inaccurate, so we re-annotated those instances.
\end{itemize}

Ultimately, we acquire 4,966 cleaner samples from Mind2Web training dataset. We also performed filtering and categorization on the GUI-R1-3K~\cite{luo2025gui} dataset, incorporating 1,744 web samples and 85 desktop samples into our high-quality dataset. As a result, we obtained a total of 6,795 high-quality data for the web and desktop domain. For the mobile platform, we leverage the AMEX~\cite{chai2025amex} dataset for its proper task difficulty, clear task descriptions and accurate  trajectory annotations. We select 3,200 data samples from AMEX for Stage 2 training.

We split the multi-step data into individual steps and trained the model using the GRPO algorithm, as described in \textbf{Appendix A}. Our reward design in Stage 2 is similar to Stage 1, except that in Stage 2, we only have a single ground truth bounding box (for both the web and desktop platform) or a ground truth point (for the mobile platform) to calculate position reward for each sample.

\section{Experiments}

\subsection{Benchmarks \& Metrics}

To comprehensively assess our GUI agent across different platforms, we evaluate it on the following six benchmarks: 

\begin{itemize}
    \item Mind2Web~\cite{deng2023mind2web} test set has three splits: Cross-Task (seen website, new task), Cross-Website (unseen website), and Cross-Domain (unseen domain). We report grounding \emph{Element Accuracy}, \emph{Operation F1} (the token-level F1 of the predicted operation), and \emph{Step Success Rate}.
    \item ScreenSpot-Pro~\cite{li2025screenspotproguigroundingprofessional} is a GUI grounding benchmark of desktop and mobile applications across six scenarios. Each target is annotated as \emph{Text} (has textual labels) or \emph{Icon} (otherwise). We report the grounding accuracy.
     \item For OmniACT~\cite{kapoor2024omniact} and AndroidControl~\cite{li2024effectsdatascaleui}, we use \emph{Type} (action type accuracy), \emph{GR} (grounding accuracy), and \emph{SR} (step success rate) as the metrics.
     \item AITW~\cite{rawles2023androidinthewild} is a large-scale dataset for mobile control that encompasses both apps and web. It has five sub domains. We report the \emph{Step Success Rate} of each domain and the overall \emph{Step Success Rate} across all domains.
     \item AndroidWorld~\cite{rawles2025androidworld} is an online benchmark that evaluates an agent’s ability to complete multi-step tasks in real Android environments. We report the \emph{Episode Success Rate} as the metric.
\end{itemize}

\subsection{Baselines}

Our primary baselines are UI-TARS~\cite{qin2025uitarspioneeringautomatedgui} and GUI-R1~\cite{luo2025gui}, and we also compare against UI-R1~\cite{lu2025uir1enhancingefficientaction}, ShowUI~\cite{Lin2025showui}, GPT-4o~\cite{gpt4oreport} and so on.

\begin{itemize}
    \item \textbf{UI-TARS}~\cite{qin2025uitarspioneeringautomatedgui}, a powerful model using large-scale SFT and Agent-DPO on 18.4 M trajectories and other data.
    \item \textbf{GUI-R1}~\cite{luo2025gui}, an R1-style RL method using a binary point-in-box reward.
    The original GUI-R1 uses about 3K training data. We also reproduce this baseline using the full mind2web training data to compare with our method.  
\end{itemize}

\subsection{Experimental Details}

We train our GUICrafter-3B model based on Qwen2.5-VL-3B~\cite{bai2025qwen25vltechnicalreport}. The training is conducted on 8 NVIDIA H20 GPUs. 
We divided our data into two platforms: the web \& desktop platform and the mobile device platform. Correspondingly, we trained two independent models. For the web \& desktop platform, we obtained 500K weakly-supervised samples. In our main experiments, we used 20,000 samples, while in the scalability study, we utilized the entire dataset. In Stage 2, we use 6,795 human-annotated web and desktop data. For the mobile platform, we obtained 136K weakly-supervised samples and used 9,600 samples in the main experiments. In Stage 2, we use 3,200 annotated mobile data.

\subsection{Main Results}

\begin{table*}[t]
\centering
\caption{Results on Mind2Web. All experiments are conducted under the same zero-shot prompt for fair comparison. The best results are in bold. The improvement brought by our Stage 1 (Weakly-Supervised GUI Pretraining) is highlighted in blue background.}
\label{tab:mind2web}
\resizebox{\linewidth}{!}{%
\begin{tabular}{lccccccccccc}
\toprule
\multirow{2}{*}{\textbf{Method}} & {\textbf{Annotated}} & \multicolumn{3}{c}{\textbf{Cross-Task}} & \multicolumn{3}{c}{\textbf{Cross-Website}} & \multicolumn{3}{c}{\textbf{Cross-Domain}} & \textbf{All} \\
\cmidrule(lr){3-5} \cmidrule(lr){6-8} \cmidrule(lr){9-11} \cmidrule{12-12}
 & {\textbf{Data Num}} & \textbf{Ele.Acc} & \textbf{Op.F1} & \textbf{Step SR} & \textbf{Ele.Acc} & \textbf{Op.F1} & \textbf{Step SR} & \textbf{Ele.Acc} & \textbf{Op.F1} & \textbf{Step SR} & \textbf{Ele.Acc} \\
\midrule
\multicolumn{10}{l}{\textbf{Agent Framework}} \\
\midrule
GPT-4o~\cite{gpt4oreport} \quad SeeClick~\cite{cheng-etal-2024-seeclick} & - & 32.1 & - & - & 33.1 & - & - & 33.5 & - & - & 33.1 \\
GPT-4V~\cite{gpt4vreport}~~~~OmniParser~\cite{lu2024omniparserpurevisionbased} & - & 42.4 & 87.6 & 39.4 & 41.0 & 84.8 & 36.5 & 45.5 & 85.7 & 42.0 & 44.1\\
GPT-4o~\cite{gpt4oreport} \quad UGround~\cite{gou2024uground} & - & 47.7 & - & - & 46.0 & - & - & 46.6 & - & - & 46.8 \\
GPT-4o~\cite{gpt4oreport} \quad Aria-UI~\cite{yang-etal-2025-aria} & - & 57.6 & - & - & 57.7 & - & - & 61.4 & - & - & 60.8 \\
\midrule
\multicolumn{10}{l}{\textbf{Proprietary Model}} \\
\midrule
GPT-4o~\cite{gpt4oreport} & - & 5.7 & 77.2 & 4.3 & 5.7 & 79.0 & 3.9 & 5.5 & 86.4 & 4.5 & 5.6\\
GPT-4 (SOM)~\cite{gpt4technicalreport} & - & 29.6 & - & 20.3 & 20.1 & - & 13.9 & 27.0 & - & 23.7 & 26.6 \\
GPT-4 (Text-only)~\cite{gpt4technicalreport} & - & 40.8 & 63.1 & 32.3 & 30.2 & 61.0 & 27.0 & 35.4 & 61.9 & 29.7 & 35.8\\
\midrule
\multicolumn{10}{l}{\textbf{Open-Source Model (About 3B Parameter Size)}} \\
\midrule
ShowUI-2B (Zero-Shot)~\cite{Lin2025showui} & 0 & 21.4 & 85.2 & 18.6 & 21.9 & 81.9 & 16.8 & 24.4 & 83.9 & 21.4 & 23.4\\
Qwen2.5-VL-3B (Zero-Shot)~\cite{bai2025qwen25vltechnicalreport} & 0 & \colorbox{blue!10}{25.5} & 33.3 & 21.1 & \colorbox{blue!10}{23.7} & 31.1 & 18.3 & \colorbox{blue!10}{28.8} & 34.1 & 22.6 & \colorbox{blue!10}{27.3}\\
GUI-R1-3B$^1$~\cite{luo2025gui} & 3.0 K &  36.0 & 34.6 & 30.7 & 39.2 & 36.4 & 31.7 & 38.9 & 35.9 & 32.0 & 38.3\\
ShowUI-2B (Fine-tuned)~\cite{Lin2025showui}  & 7.7 K & 39.9 & 88.6 & 37.2 & 41.6 & 83.5 & 35.1 & 39.4 & 86.8 & 35.2 & 39.8 \\
UI-R1-3B~\cite{lu2025uir1enhancingefficientaction} & 3.0 K & 34.4 & - & - & 41.1 & - & - & 42.5 & - & - & 40.5\\
Qwen2.5-VL-3B (Fine-tuned)~\cite{bai2025qwen25vltechnicalreport} & 7.7 K & 51.4 & 65.4 & 47.7 & 47.6 & 53.3 & 41.5 & 47.0 & 55.7 & 43.2 & 48.1\\
GUI-R1-3B$^2$~\cite{luo2025gui}  & 7.7 K & 53.9 & 87.1 & 48.9 & 55.6 & 84.0 & 48.1 & 54.7 & 85.8 & 49.8 & 54.7\\
UI-TARS-2B~\cite{qin2025uitarspioneeringautomatedgui} & 18.4 M & \textbf{62.3} & \textbf{90.0} & \textbf{56.3} & 58.5 & 87.2 & 50.8 & 58.8 & 89.6 & 52.3 & 59.5 \\
\midrule
GUICrafter-3B (ours) & 6.7 K & 59.4 & 89.3 & 54.0 & \textbf{59.2} & \textbf{87.6} & \textbf{54.1} & \textbf{60.7} & \textbf{90.6} & \textbf{52.9} & \textbf{60.2}\\
\quad -- only Stage1 & 0 & \colorbox{blue!10}{37.2 (11.7 $\uparrow$)} & 37.2 & 33.8 & \colorbox{blue!10}{40.5 (16.8 $\uparrow$)} & 39.2 & 36.3 & \colorbox{blue!10}{40.4 (11.6 $\uparrow$)} & 37.5 & 36.7 & \colorbox{blue!10}{39.7 (12.4 $\uparrow$)}\\
\quad -- only Stage2 & 6.7 K & 56.9 & 87.7  & 51.8  & 56.3 & 84.9  & 48.9  & 56.2 & 86.7  & 50.7 & 56.4\\
\midrule
\multicolumn{10}{l}{\textbf{Open-Source Model (About 7B Parameter Size)}} \\
\midrule
Qwen2.5-VL-7B (Zero-Shot)~\cite{bai2025qwen25vltechnicalreport} & 0 & \colorbox{blue!10}{34.0} & 36.3 & 25.6 & \colorbox{blue!10}{35.1} & 38.4 & 23.6 & \colorbox{blue!10}{36.4} & 40.4 & 25.3 & \colorbox{blue!10}{35.7}\\
GUI-R1-7B$^1$~\cite{luo2025gui} & 3.0 K & 41.4 & 42.7 & 33.8 & 44.5 & 44.7 & 35.0 & 44.6 & 47.0 & 35.7 & 43.9 \\
Qwen2.5-VL-7B (Fine-tuned)~\cite{bai2025qwen25vltechnicalreport} & 7.7 K & 65.1 & 90.0 & 60.8 & 59.9 & 87.8 & 53.9 & 60.2 & 89.2 & 56.4 & 61.3\\
Aguvis-7B~\cite{xu2024aguvis} & 5.5 M & 64.2 & 89.8 & 60.4 & 60.7 & 88.1 & 54.6 & 60.4 & 89.2 & 56.6 & 61.3\\
GUI-R1-7B$^2$~\cite{luo2025gui}  & 7.7 K & 73.0 & 91.2 & 66.8 & 67.6 & 89.1 & 60.0 & 65.9 & 88.6 & 59.0 & 67.7\\
UI-TARS-7B~\cite{qin2025uitarspioneeringautomatedgui} & 18.4 M & 73.1 & 92.2 & 67.1 & 68.2 & 90.9 & 61.7 & 66.6 & 90.9 & 60.5 & 68.3\\
\midrule
GUICrafter-7B (ours) & 6.7 K & \textbf{74.4} & \textbf{92.5} & \textbf{68.4} & \textbf{70.2} & \textbf{91.5} & \textbf{63.0} & \textbf{68.4} & \textbf{91.3} & \textbf{62.2} & \textbf{70.0} \\
\quad -- only Stage1 & 0 & \colorbox{blue!10}{51.8 (17.8 $\uparrow$)} & 64.7 & 45.9 & \colorbox{blue!10}{50.8 (15.7 $\uparrow$)} & 55.7 & 44.9 & \colorbox{blue!10}{51.0 (14.6 $\uparrow$)} & 58.8 & 46.1 & \colorbox{blue!10}{51.1 (15.4 $\uparrow$)} \\
\quad -- only Stage2 & 6.7 K & 72.1 & 90.9 & 66.4 & 68.3 & 89.8 & 61.1 & 66.7 & 89.7 &  60.1 & 68.1\\

\bottomrule
\addlinespace[0.5em]
\multicolumn{10}{l}{$^1$ Originally, GUI-R1 mannually select 100 data from Mind2Web, along with other high-quality data, to train the model.} \\
\multicolumn{10}{l}{$^2$ GUI-R1 trained on full Mind2Web training dataset.} \\
\end{tabular}%
}
\end{table*}

\begin{table*}[t]
\centering
\caption{GUI grounding accuracy on ScreenSpot-Pro. All experiments are conducted under the same zero-shot prompt for fair comparison. * denotes supervised fine-tuned on GUI-R1-3K~\cite{luo2025gui}. The best results are in bold. The improvement brought by our Stage 1 (Weakly-Supervised GUI Pretraining) is highlighted in blue background.}
\label{tab:screenspot_pro}
\resizebox{\textwidth}{!}{
\begin{tabular}{lccccccccccccccccccc}
\toprule
\multirow{2}{*}{\textbf{Models}} & \multicolumn{3}{c}{Dev} & \multicolumn{3}{c}{Creative} & \multicolumn{3}{c}{CAD} & \multicolumn{3}{c}{Scientific} & \multicolumn{3}{c}{Office} & \multicolumn{3}{c}{OS} & All  \\
\cmidrule(lr){2-4} \cmidrule(lr){5-7} \cmidrule(lr){8-10} \cmidrule(lr){11-13} \cmidrule(lr){14-16} \cmidrule(lr){17-19} \cmidrule(lr){20-20}
& Text & Icon & Avg. & Text & Icon & Avg. & Text & Icon & Avg. & Text & Icon & Avg. & Text & Icon & Avg. & Text & Icon & Avg. & Avg. \\
\midrule
\multicolumn{20}{l}{\textbf{Proprietary Model}} \\
\midrule
GPT-4o~\cite{gpt4oreport} & 1.3 & 0.0 &0.7& 1.0 & 0.0 &0.6& 2.0 & 0.0 &1.5& 2.1 & 0.0 &1.2& 1.1 & 0.0 &0.8& 0.0 & 0.0 & 0.0 & 0.8\\
Claude CUA~\cite{claudeCUA} & 22.0 & 3.9 &13.2& 25.9 & 3.4 &16.5& 14.5 & 3.7 &11.9& 33.9 & 15.8 &26.1& 30.1 & 16.3 &26.9& 11.0 & 4.5 &8.0&17.1\\
\midrule
\multicolumn{20}{l}{\textbf{Open-Source Model (About 3B Parameter Size)}} \\
\midrule
OS-Atlas-4B~\cite{wu2024atlas} & 7.1 & 0.0 & 3.7 & 3.0 & 1.4 & 2.3 & 2.0 & 0.0 & 1.5& 9.0 & 5.5& 7.5 & 5.1 & 3.8 & 4.8& 5.6 & 0.0 & 3.1 & 3.7\\
ShowUI-2B~\cite{Lin2025showui} & 16.9 & 1.4 & 9.4 & 9.1 & 0.0 & 5.3 & 2.5 & 0.0 & 1.9& 13.2 & 7.3 & 10.6& 15.3 & 7.5& 13.5 & 10.3 & 2.2 & 6.6 & 7.7 \\
Qwen2.5-VL-3B~\cite{bai2025qwen25vltechnicalreport} & 16.2 & 1.4 & 9.0& 23.3 & 1.4 & 14.1 & 10.2 & 4.7 & 8.9 & 38.2 & 6.4 & 24.4 & 24.3 & 3.8 & 19.6 & 15.0 & 1.1 & 8.7 & \colorbox{blue!10}{14.1} \\
Qwen2.5-VL-3B*~\cite{bai2025qwen25vltechnicalreport} & 20.3 & 1.8 &11.3& 24.6 & 2.8 &15.5& 11.2 & 4.7 &9.6& 39.5 & 6.4 &25.2& 28.6 & 5.7 &23.3& 17.8 & 2.2&10.7&15.8\\
UI-R1-3B~\cite{lu2025uir1enhancingefficientaction} & 22.7 & 4.1 & 13.7& 27.3 & 3.5 & 17.3& 11.2 & 6.3 & 10.0& 43.4 & 11.8& 29.7& 32.2 & 11.3 &27.4& 13.1 & 4.5 &9.2&17.9\\
UI-TARS-2B~\cite{qin2025uitarspioneeringautomatedgui} & \textbf{47.4} & 4.1 & \textbf{26.4} & \textbf{42.9} & \textbf{6.3} & \textbf{27.6} & 17.8 & 4.7 & 14.6& 56.9 & 17.3 & 39.8& 50.3 & 17.0 & 42.6& 21.5 & 5.6 & 14.3& 27.7\\
GUI-R1-3B~\cite{luo2025gui} & 33.8 & 4.8 & 19.7 & 40.9 & 5.6 & 26.1 & 26.4 & 7.8 & 21.8 & \textbf{61.8} & 17.3 & 42.5 & 53.6 & 17.0 & 45.2 & 28.1 & 5.6 & 17.9 & 28.6\\
\midrule
GUICrafter-3B (ours) & 40.3 & \textbf{6.2} & 23.8 & \textbf{42.9} & \textbf{6.3} & \textbf{27.6} & \textbf{41.6} & \textbf{10.9} & \textbf{34.1} & 55.6& \textbf{20.9} & 40.6 & \textbf{66.1} & \textbf{20.8} & \textbf{55.7} & \textbf{33.6} & \textbf{9.0} & \textbf{22.4} & \textbf{33.5}\\
\quad -- only Stage1 & 28.6 & 3.4 & 16.4 & 38.9 & 4.9 & 24.6 & 24.9 & 4.7 & 19.9 & 50.7 & 12.7 & 34.3 & 42.4 & 11.3 & 35.2 & 26.2 & 5.6 & 16.8& \colorbox{blue!10}{24.4 (10.3 $\uparrow$)}\\
\quad -- only Stage2 & 35.7 & 4.1 & 20.4 & 40.9 & 5.6 & 26.1 & 36.0 & 7.8 & 29.1& 60.4 & 20.0 & \textbf{42.9} & 55.4 & 13.2 & 45.7 & 32.7 & \textbf{9.0} & 21.9 & 30.5\\
\midrule
\multicolumn{20}{l}{\textbf{Open-Source Model (About 7B Parameter Size)}} \\
\midrule
SeeClick-9.6B~\cite{cheng-etal-2024-seeclick} & 0.6 & 0.0 & 0.3 & 1.0 & 0.0 &0.6 & 2.5 & 0.0 & 1.9 & 3.5 & 0.0 & 2.0 & 1.1 & 0.0& 0.8 & 2.8 & 0.0 & 1.5 & 1.1\\
CogAgent-18B~\cite{Hong_2024_CVPR} & 14.9 & 0.7 & 8.0 & 9.6 & 0.0 & 5.6& 7.1 & 3.1 &6.1& 22.2 & 1.8 &13.4& 13.0 & 0.0 &10.0& 5.6 & 0.0 &3.1&7.7\\
Aria-UI~\cite{yang-etal-2025-aria} & 16.2 & 0.0 & 8.3 & 23.7 & 2.1 &14.6& 7.6 & 1.6 &6.1& 27.1 & 6.4 &18.1& 20.3 & 1.9 &16.1& 4.7 & 0.0&2.6&11.3\\
UGround-7B~\cite{gou2024uground} & 26.6 & 2.1 &14.7& 27.3 & 2.8 &17.0& 14.2 & 1.6 &11.1& 31.9 & 2.7 &19.3& 31.6 & 1.3 &24.6& 17.8 & 0.0&9.7&16.2 \\
OS-Atlas-7B~\cite{wu2024atlas} & 33.1 & 1.4 & 17.7& 28.8 & 2.8 &17.9& 12.2 & 4.7 &10.4& 37.5 & 7.3 &24.4& 33.9 & 5.7 &27.4& 27.1 & 4.5&16.8&18.9 \\
Qwen2.5-VL-7B~\cite{bai2025qwen25vltechnicalreport} & 33.1 & 2.1 & 18.1 & 23.7 & 3.5 & 15.2 & 12.2 & 6.3 & 10.8& 36.8 & 7.3 &24.0& 37.8 & 7.5 &30.8& 30.8 & 6.9&19.9& \colorbox{blue!10}{19.3} \\
Qwen2.5-VL-7B*~\cite{bai2025qwen25vltechnicalreport} & 31.4 & 1.8& 17.0& 27.3 & 3.5 &17.3& 15.7 & 5.1 &13.1& 40.7 & 7.9 &26.5& 39.7 & 8.9 &32.6& 32.4 & 6.9&20.8&20.7\\
GUI-R1-7B~\cite{luo2025gui} & 49.4 & 4.8 & 27.8 & 38.9 & 8.4 & 26.1 & 23.9 & 6.3 & 19.6 & 55.6 & 11.8 & 36.6 & 58.7 & 26.4 & 51.3 & 42.1 & 16.9 7 & 30.7 & 31.3 \\
UI-TARS-7B & \textbf{58.4} & 12.4 & 36.1 & 50.0 & 9.1 & 32.8 & 20.8 & 9.4 & 18.0 & \textbf{63.9} & 31.8 & \textbf{50.0} & 63.3 & 20.8 & 53.5 & 30.8 & 16.9 & 24.5 & 35.7 \\
\midrule

GUICrafter-7B (ours) 
& 58.0 & \textbf{14.2} & \textbf{36.8} 
& \textbf{54.3} & \textbf{9.8} & \textbf{35.6}
& \textbf{35.4} & \textbf{11.0} & \textbf{29.4} 
& 62.8 & \textbf{32.5} & 49.7
& \textbf{69.5} & \textbf{25.9} & \textbf{59.5}
& \textbf{33.9} & \textbf{18.6} & \textbf{27.0}
& \textbf{39.5} \\

\quad -- only Stage1 
& 48.0 & 9.8 & 29.5
& 39.6 & 7.1 & 26.0
& 18.6 & 8.0 & 16.0
& 50.9 & 24.5 & 39.5
& 53.7 & 18.9 & 45.7
& 24.7 & 13.6 & 19.7
& \colorbox{blue!10}{29.2 (9.9 $\uparrow$)}\\

\quad -- only Stage2 
& 57.3 & 12.9 & 35.8
& 49.4 & 8.9 & 32.4
& 23.1 & 10.0 & 19.9
& 63.5 & 30.5 & 49.2 
& 63.2 & 23.6 & 54.1
& 30.8 & 16.9 & 24.5
& 35.8\\

\bottomrule
\end{tabular}
}
\end{table*}

\paragraph{\textbf{Results on Mind2Web}}

The results of GUICrafter on Mind2Web~\cite{deng2023mind2web} are presented in Table \ref{tab:mind2web}. Based on these results, we have the following five conclusions:

\begin{itemize}
    \item After only Stage 1 (Weakly-Supervised GUI Pretraining), our model achieves an average accuracy improvement of over 10\% across all subcategories of Mind2Web, compared to the base model Qwen2.5-VL-3B. This entire process requires no human labor.
    \item After Stage 1 and Stage 2 (High-Quality Reinforcement Fine-Tuning), our GUICrafter model achieves the best performance of average grounding accuracy. It is worth noting that UI-TARS is trained on approximately 18.4M samples, whereas our training data include only 6,795 high-quality samples and 20,000 weakly-supervised samples collected without any human annotation cost. Therefore, our method achieves results comparable to UI-TARS using only about one-thousandth of its data, demonstrating the high data efficiency of our approach. 
    \item Our model outperforms UI-TARS on the cross-website and cross-domain subsets more significantly, which exhibit larger distributional differences from the training set. This demonstrates that our method provides stronger generalization ability, primarily benefiting from Stage 1, where the model is exposed to a large number of real-world webpages.
    \item The curriculum learning process that includes both Stage 1 and Stage 2 leads to a 3\%–4\% improvement in grounding accuracy compared to training with Stage 2 alone, clearly demonstrating that GUI Pretraining provides a solid foundation for the model.
    \item We also reproduced GUI-R1 using the entire Mind2Web training set for a fair comparison. Our model outperforms this reproduced GUI-R1 baseline significantly.
\end{itemize}

\paragraph{\textbf{Results on ScreenSpot-Pro}}

The results of GUICrafter on ScreenSpot-Pro~\cite{li2025screenspotproguigroundingprofessional} are presented in Table \ref{tab:screenspot_pro}. Based on these results, we have the following three conclusions:


\begin{itemize}[topsep=2pt, itemsep=2pt, parsep=0pt, partopsep=0pt]
    \item After only Stage 1 (Weakly-Supervised GUI Pretraining), our model achieves about 10\% improvement in average accuracy over the base model Qwen2.5-VL-3B, with no human labor required throughout this process. 
    \item After Stage 1 and Stage 2 (High-Quality Reinforcement Fine-Tuning), our model achieves the best performance on ScreenSpot-Pro among models of comparable size. Specifically, our model surpasses the second-best model, GUI-R1-3B, by 4\%-5\% in average accuracy, and outperforms it in the majority of subcategories.
    \item The curriculum learning process that includes both Stage 1 and Stage 2 yields a 3\% improvement over training with Stage 2 alone, which provides clear evidence that GUI pretraining offers a reliable basis for the model.
\end{itemize}

\paragraph{\textbf{Results on AndroidControl and AITW}}

The results of GUICrafter on AndroidControl~\cite{li2024effectsdatascaleui} and AITW~\cite{rawles2023androidinthewild} are presented in Tables \ref{tab:android_control} and \ref{tab:aitw}. In Table \ref{tab:android_control}, * denotes supervised fine-tuned on GUI-R1-3K~\cite{luo2025gui}. We adopt a zero-shot setting when evaluating our model on AITW. Based on these results, we have the following three conclusions:
\begin{itemize}
    \item After Stage 1, GUICrafter achieves 62.35\% Step SR on AndroidControl-Low and 44.65\% on AndroidControl-High without using any annotated data, which are comparable to GUI-R1-3B~\cite{luo2025gui} trained on human-annotated data.
    \item After both Stage 1 and Stage 2, GUICrafter outperforms other models of comparable size on the AndroidControl benchmark. Without fine-tuning on the training dataset of AITW, GUICrafter demonstrates generalization capability and outperforms other zero-shot models on AITW.
    \item After the whole curriculum learning process, GUICrafter, using merely 3B parameters and 3,200 human-annotated data, yields an overall zero-shot performance of 50.89\% on AITW, which is close to the performance of methods relying on close-source models with other assistance like GPT-4V + history~\cite{yan2023gpt} and OmniParser~\cite{lu2024omniparserpurevisionbased}.
\end{itemize}


\begin{table*}[t]
\centering

\begin{minipage}[t]{0.48\textwidth}
\centering
\caption{Results on AndroidControl.}
\label{tab:android_control}
\resizebox{\linewidth}{!}{
\begin{tabular}{lcccccc}
\toprule
\multirow{2}{*}{\textbf{Models}} & 
\multicolumn{3}{c}{\textbf{AndroidControl-Low}} & 
\multicolumn{3}{c}{\textbf{AndroidControl-High}} \\
\cmidrule(lr){2-4} \cmidrule(lr){5-7}
 & Type & GR & SR & Type & GR & SR \\
\midrule
GPT-4o~\cite{gpt4oreport}        & 74.33 & 38.67 & 28.39 & 63.06 & 30.90 & 21.17 \\
Qwen2.5-VL-3B~\cite{bai2025qwen25vltechnicalreport}  & 62.03 & \colorbox{blue!10}{74.07} & 59.32 & 47.81 & \colorbox{blue!10}{46.51} & 38.90 \\
Qwen2.5-VL-7B~\cite{bai2025qwen25vltechnicalreport}  & 83.44 & 87.08 & 62.50 & 68.67 & 59.71 & 47.06 \\
OS-Atlas-7B~\cite{wu2024atlas}   & 73.00 & 73.37 & 50.94 & 57.44 & 54.90 & 29.83 \\
Qwen2.5-VL-3B*~\cite{bai2025qwen25vltechnicalreport} & 71.08 & 74.53 & 58.79 & 52.05 & 49.53 & 41.22 \\
Qwen2.5-VL-7B*~\cite{bai2025qwen25vltechnicalreport} & 84.00 & 85.74 & 64.32 & 69.15 & 58.69 & 48.11 \\
UI-R1-3B~\cite{lu2025uir1enhancingefficientaction}      & 79.15 & 82.41 & 66.44 & 57.85 & 55.70 & 45.44 \\
GUI-R1-3B~\cite{luo2025gui}     & 83.68 & 81.59 & 64.41 & 58.04 & 56.24 & 46.55 \\
\midrule
GUICrafter--3B (ours)     & 86.79 & \textbf{82.08} & 70.73 & \textbf{73.38} & \textbf{58.96} & \textbf{56.50} \\
\quad -- only Stage1 & 82.72 & \colorbox{blue!10}{81.39 (7.32 $\uparrow$)} & 62.35 & 69.64 & \colorbox{blue!10}{57.57 (11.06 $\uparrow$)} & 44.65 \\
\quad -- only Stage2     & 86.14 & 78.37 & 70.16 & 71.73 & 58.18 & 54.17 \\
\quad -- Binary Reward     & \textbf{86.92} & 79.78 & \textbf{71.05} & 72.44 & 57.41 & 54.89 \\
\bottomrule
\end{tabular}
}
\end{minipage}
\hfill
\begin{minipage}[t]{0.48\textwidth}
\centering
\caption{Zero-shot results on AITW.}
\label{tab:aitw}
\resizebox{\linewidth}{!}{
\begin{tabular}{lcccccc}
\toprule
\textbf{Method} & \textbf{General} & \textbf{Install} & \textbf{G.Apps} & \textbf{Single} & \textbf{WebShop.} & \textbf{Overall} \\
\midrule
\multicolumn{7}{l}{\textbf{Close-Source Model}} \\
\midrule
ChatGPT-CoT~\cite{zhang2024you} & 5.93 & 4.38 & 10.47 & 9.39 & 8.42 & 7.72 \\
PalM2-CoT~\cite{rawles2023androidinthewild} & - & - & - & - & - & 39.6 \\
GPT-4V + history~\cite{yan2023gpt} & 43.01 & 46.14 & 49.18 & 78.29 & 48.18 & 52.96 \\
OmniParser (w. LS + ID)~\cite{lu2024omniparserpurevisionbased} & 48.3 & 57.8 & 51.6 & 77.4 & 52.9 & 57.7 \\
\midrule
\multicolumn{7}{l}{\textbf{Open-Source Model}} \\
\midrule
Qwen2.5-VL-3B~\cite{bai2025qwen25vltechnicalreport} & 27.57 & 28.58 & 22.97 & 37.28 & 24.30 & \colorbox{blue!10}{28.14} \\
ShowUI-2B~\cite{Lin2025showui} & 32.1 & 47.7 & 42.0 & 20.1 & 37.4 & 35.9 \\
GUI-R1-3B~\cite{luo2025gui} & 43.48 & 44.87 & 39.97 & 54.06 & 35.60 & 43.60 \\
\midrule
GUICrafter--3B (ours) & \textbf{46.57} & \textbf{54.29} & 51.86 & \textbf{61.46} & 40.25 & \textbf{50.89} \\
\quad -- only Stage1 & 30.09 & 34.07 & 32.73 & 36.26 & 24.50 & \colorbox{blue!10}{31.53 (3.39 $\uparrow$)} \\
\quad -- only Stage2 & 45.00 & 52.11 & \textbf{52.36} & 58.84 & \textbf{41.28} & 49.92 \\
\quad -- Binary Reward & 44.44 & 52.28 & 51.15 & 57.68 & 36.99 & 48.51 \\
\bottomrule
\end{tabular}
}
\end{minipage}

\end{table*}

\paragraph{\textbf{Results on OmniACT}}
We present results on OmniACT in \textbf{Appendix B}. GUICrafter also achieves leading performance on the OmniACT benchmark.

\paragraph{\textbf{Results on AndroidWorld}}

\begin{wraptable}{r}{0.3\columnwidth}
\vspace{-10pt}
\centering
\caption{Results on AndroidWorld.}
\label{tab:androidworld}
\resizebox{\linewidth}{!}{
\begin{tabular}{lc}
\toprule
\textbf{Models} & \textbf{Episode SR} \\
\midrule
Qwen2.5-VL-3B~\cite{bai2025qwen25vltechnicalreport} & 10.34 \\
GUI-R1-3B~\cite{luo2025gui} & 14.22 \\
GUICrafter-3B & \textbf{25.43} \\
-- only Stage1 & 14.66 \\
\bottomrule
\end{tabular}
}
\end{wraptable}

The results of GUICrafter on AndroidWorld are presented in Table \ref{tab:androidworld}. Table \ref{tab:androidworld} demonstrates that GUICrafter also performs strongly on online, end-to-end benchmarks such as AndroidWorld. With only Stage 1 training, it already achieves performance comparable to GUI-R1; after the full Stage 1+2 training, it surpasses GUI-R1 by approximately 11\% in Episode Success Rate.






\subsection{Ablation Studies}
\paragraph{\textbf{Ablation on Weakly-Supervised GUI Pretrain}} \quad We investigate the impact of Stage 1 (Weakly-Supervised GUI Pretraining) by conducting experiments where the model undergoes only Stage 2 (High-Quality Reinforcement Fine-Tuning) and comparing them with the GUICrafter full version. The results of these experiments are reported in the main result tables for each benchmark (Tables \ref{tab:mind2web}, \ref{tab:screenspot_pro}, \ref{tab:android_control}, \ref{tab:aitw}).
We find that the combination of Stage 1 and Stage 2 significantly outperforms training with Stage 2 alone — for example, achieving 4\% improvement on Mind2Web and 3\% improvement on ScreenSpot. This indicates that our Stage 1 GUI Pretraining expands the model’s capability boundaries. Even though the weakly-supervised signals may contain noise, the model effectively learns from them and internalizes this knowledge as GUI-specific understanding.
\paragraph{\textbf{Ablation on Gaussian Reward}}

\quad To showcase the necessity of introducing the Gaussian position reward in Stage 1, we conduct an ablation study by modifying the reward design to a simple binary reward, i.e., the position reward is set to 1 if the predicted point falls within any interactive region of the screen, otherwise 0. The results are reported in the main results tables for AndroidControl and AITW (Tables \ref{tab:android_control} and \ref{tab:aitw}). We found that the Gaussian reward outperforms the binary reward on all metrics on AndroidControl-High and AITW, 
and by 2.3\% on grounding accuracy on AndroidControl-Low. This demonstrates the negative impact if we ignore the center positions of interactive regions, clarifying the need for the Gaussian reward.
\paragraph{\textbf{Ablation on Task Formulation}}

\quad To assess whether the abstraction into meta-tasks limits the agent’s capability, we conduct an ablation study on different task formulations.
(1) \textbf{only-click task}, a highly constrained prompt that only implies a click action (e.g., \textit{Click any clickable area on the page, such as a button, but not a blank space}); (2) \textbf{meta-task} used in the paper, covering all action types in the unified action space; (3) \textbf{LLM-gen task}, automatically generated by GPT-4o~\cite{gpt4oreport} API, where a GUI element is randomly selected from the screenshot and a corresponding task prompt is synthesized; and (4) \textbf{human-annotated task}. In Stage 2, we still fine-tune the model using the same small amount of high-quality data. 
We train separate models with each task formulation while keeping all other factors fixed, using the same number of training data as the original paper. 

For evaluation, we construct a complex subset from Mind2Web~\cite{deng2023mind2web}. 
In terms of difficulty, we select test episodes from unseen domains. In terms of trajectory length, we select episodes with more than 10 steps. 
As a result, we obtain a hard subset consisting of 148 episodes, with an average length of 13.57 steps.

The evaluation results in Table \ref{tab:task_comparison} show that, 
(1) The step success rate of the click task is low in both stages, because the model degenerates into always predicting click actions. 
(2) In Stage 1, the meta-task and LLM-generated task achieve comparable performance, both of which are inferior to the human-annotated task; however, 
after Stage 2 fine-tuning, the performance of meta-task, LLM-gen task and annotated task is similar.
This is because base models already encode rich semantic knowledge, and only need Stage 1 to acquire large-scale GUI-specific knowledge and Stage 2 with a small amount of data to calibrate. Therefore, using more semantically complex tasks in Stage 1 brings limited additional benefit. 
In summary, the meta-task is already sufficiently expressive. 
With two-stage training, this simplification has almost no impact on the model’s performance on long and complex tasks, while requiring no human annotation or LLM APIs.

\begin{table}[t]
\centering
\caption{Ablation on Task Formulation}
\vspace{-1em}
\label{tab:task_comparison}
\resizebox{\linewidth}{!}{
\begin{tabular}{lcccccccc}
\toprule
 & \multicolumn{2}{c}{only-click task} 
 & \multicolumn{2}{c}{meta-task} 
 & \multicolumn{2}{c}{LLM-gen task} 
 & \multicolumn{2}{c}{annotated task} \\
\cmidrule(lr){2-3} \cmidrule(lr){4-5} \cmidrule(lr){6-7} \cmidrule(lr){8-9}
 & Ele.Acc & Step SR 
 & Ele.Acc & Step SR
 & Ele.Acc & Step SR
 & Ele.Acc & Step SR \\
\midrule
Stage 1   & 38.4 & 27.7 & 38.5 & 34.4 & 37.9 & 34.6 & 58.2 & 51.9 \\
Stage 1+2 & 58.5 & 41.2 & 58.7 & 51.3 & 58.6 & 51.5 & 59.0 & 52.1 \\
\bottomrule
\end{tabular}
}
\end{table}

\section{Analysis}

\subsection{Data Visualization and Failure Case}

\begin{figure}[t]
  \centering
  \includegraphics[width=\linewidth]{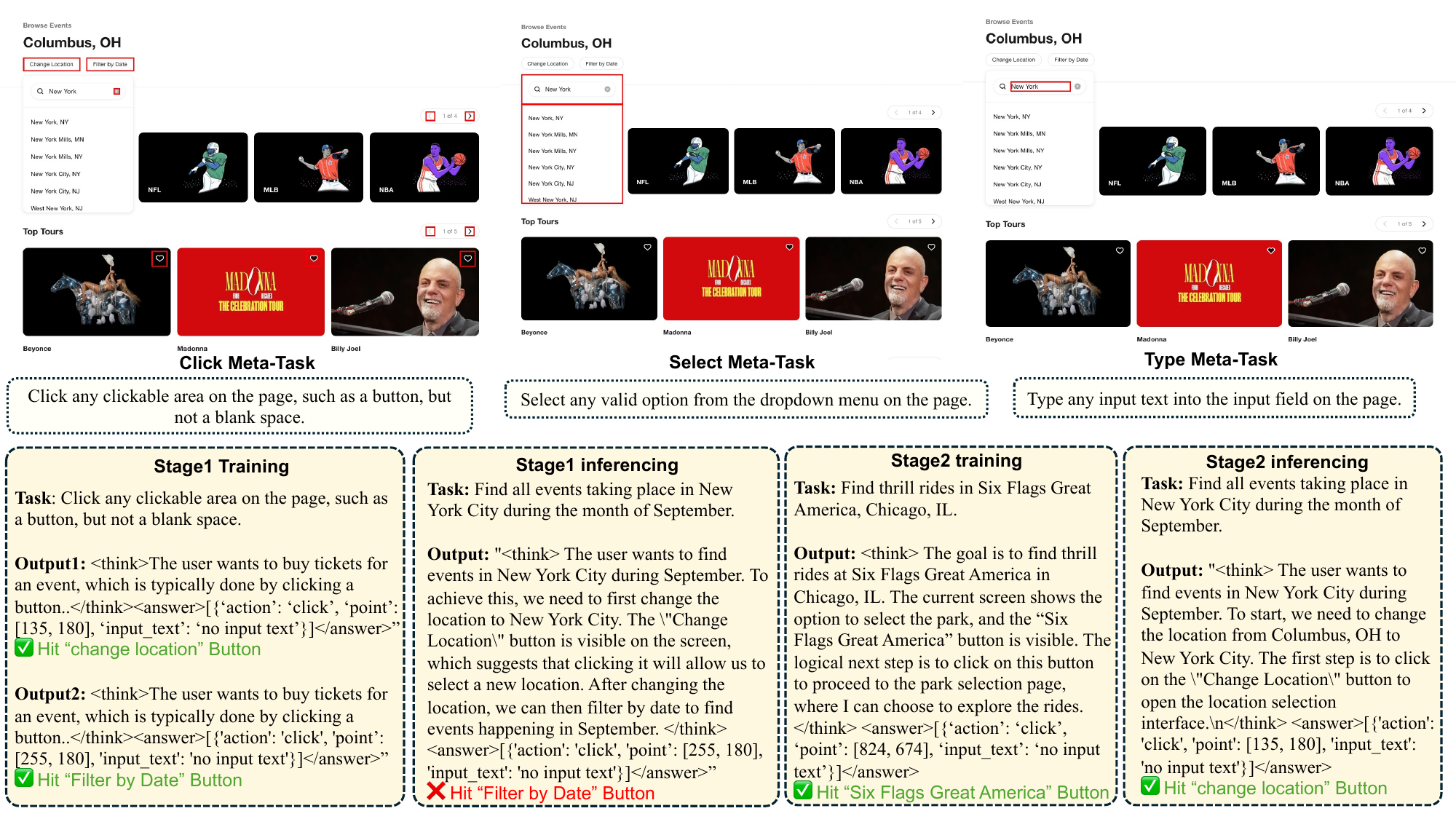}
   \caption{The top part shows raw screenshots, meta-tasks and extracted signals highlighted in red. The bottom shows the thoughts and actions at different stages. 
   }
   \label{fig:data}
   \vspace{-5pt}
\end{figure}

In Figure \ref{fig:data}, we provide Stage 1 data visualization. To clearly illustrate the effects of Stage 1 and Stage 2 training, we further present a case in which the model fails after Stage 1 training but succeeds after Stage 2 training. The top part of Figure \ref{fig:data} shows that, for the same screenshot, we define three different meta-tasks with different target region highlighted in red. The bottom part presents example tasks (which are wrapped into the prompt template and used as input to the GUI agent) and the corresponding model outputs at different stages. The webpage screenshots used for Stage 1 training, Stage 1 inferencing, and Stage 2 inferencing are the same as the screenshot displayed in the top part of the figure. In Stage 1 inferencing, the agent fails despite correct meta-task supervision from Stage 1 training. After Stage 2 training, the agent is able to solve this test case. This is because the Stage 1 data does not contain semantic information about the interactive regions; it merely distinguishes interactive regions from non-interactive ones. Stage 2, in contrast, teaches the model to identify which interactive region is the correct one.

\subsection{Data Scalability and Efficiency}

We conduct an ablation study on the amount of weakly-supervised data in Stage 1. We set the dataset to 10, 100, 1000, 10000, and 50000 samples, independently run each experiment three times, and calculate the average results, which are presented in Figure \ref{fig:scale}. We have two key conclusions:

\begin{figure}[t]
  \centering 
  \scriptsize
  \begin{overpic}[width=0.8\linewidth]{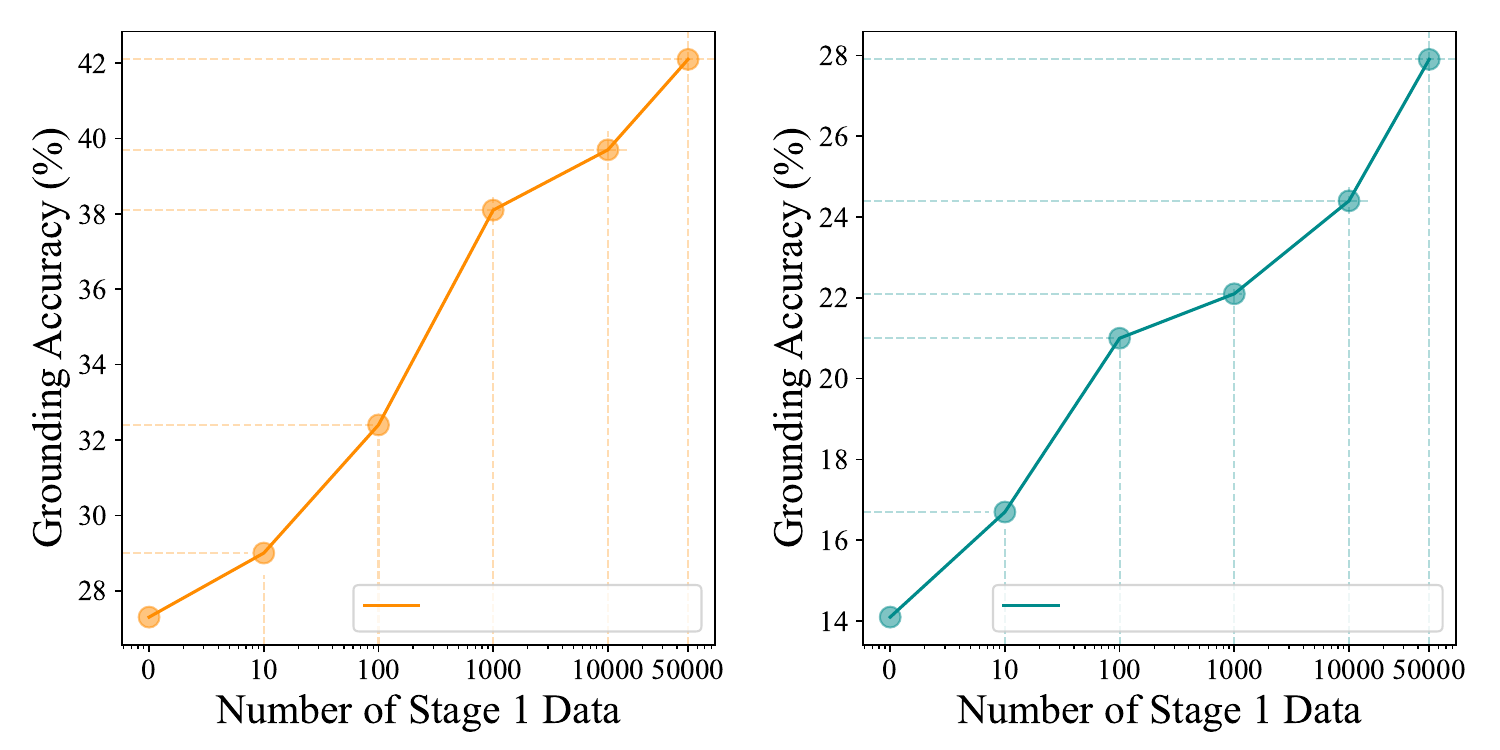}
    \put(28.5, 8.5){Mind2Web~\cite{deng2023mind2web}}
    \put(71.5, 8.5){ScreenSpot-Pro~\cite{li2025screenspotproguigroundingprofessional}}
  \end{overpic}
\caption{As the amount of Stage 1 data increases, the model’s grounding accuracy on both Mind2Web and ScreenSpot-Pro benchmarks consistently improves.}
\label{fig:scale}
\vspace{-5pt}
\end{figure}

\begin{itemize}
    \item \textbf{Data Efficiency}: Even a small amount of weakly-supervised data can enhance the agent model’s visual grounding ability. For example, using only 10 weakly-supervised samples improves performance over the base model (Qwen2.5-VL-3B) by 1.7\% on Mind2Web and 2.6\% on ScreenSpot-Pro. This clearly demonstrates the effectiveness and data efficiency of our weakly-supervised data construction.
    \item \textbf{Data Scalability}: As the data volume increases, the model continues to gain performance improvements, with no saturation observed up to the 50k data scale. This indicates that our weakly-supervised data possess scalability—even though they contain some noise, they can still scale the model’s capability effectively. With more Stage 1 data, model performance converges at about 350k Stage 1 data samples, suggesting a relatively high noise ceiling.
\end{itemize}

\subsection{Noise Analysis and Noise Robustness}

We manually inspected 1,000 randomly sampled Stage 1 data and found that 84.9\% of the samples are fully correct, without missing, overlapping or disordered visual interactable elements. In this analysis, we define noise as samples that contain missing, overlapping, or disordered visual interactable elements. To further study the robustness of the training pipeline to such imperfections, we keep the total amount of data fixed and manually adjust the noise ratio in the Stage 1 dataset by controlling the proportion of these noisy samples. We then evaluate the model's overall grounding accuracy on Mind2Web~\cite{deng2023mind2web} and ScreenSpot-Pro~\cite{li2025screenspotproguigroundingprofessional} after completing Stage 1 and Stage 2 training.

The results are summarized in Table \ref{tab:noise_robustness}. We observe that as the noise ratio increases, the performance of the model trained with only Stage 1 degrades. However, after Stage 2 training, the performance gap under different noise levels is significantly reduced. This suggests that the proposed two-stage training framework is robust to noisy supervision in the Stage 1 data.


\begin{table}[t]
\centering
\caption{Model Performance of Different Noise Ratios}
\label{tab:noise_robustness}
\begin{tabular}{lccc}
\toprule
Noise Ratio & Stage & Mind2Web & ScreenSpot-Pro \\
\midrule
\multirow{2}{*}{0\%} 
    & 1   & 40.5 & 24.4 \\
    & 1+2 & \textbf{59.1} & \textbf{32.9} \\
\midrule
\multirow{2}{*}{15\%}
    & 1   & 38.1 & 22.0 \\
    & 1+2 & 58.9 & 32.6 \\
\midrule
\multirow{2}{*}{30\%}
    & 1   & 36.9 & 19.4 \\
    & 1+2 & 58.3 & 31.8 \\
\bottomrule
\end{tabular}
\end{table}

\section{Conclusion}

This work addresses the data scarcity predicament in GUI agents and introduces GUICrafter, which
substantially reduces dependence on costly manual annotations and enables GUI agents to evolve by observing massive unannotated screenshots.
Experimental results demonstrate that GUICrafter achieves competitive or even superior performance compared with advanced systems such as UI-TARS and GUI-R1, highlighting its strong data efficiency and domain generalization ability. 
The limitation of our approach is reliance on a small amount of human-annotated data in Stage 2. In future work, we plan to explore LLM-driven reverse GUI task synthesis from GUI elements and interaction signals, aiming to establish a data flywheel that enables GUI agent's continuous self-evolution and ultimately resolve the data bottleneck in GUI agent's development.

\clearpage

%
%
\bibliographystyle{splncs04}
\bibliography{main}

\clearpage

\appendix

\begin{center}
  {\LARGE\bfseries Appendix\par}
\end{center}
\vspace{1em}

\section{Preliminaries}

\subsection{GUI Agent Formulation}

For a single-step case, the GUI agent takes input containing the task instruction $\mathcal{I}$, current observation $o$ (usually screenshot), and action history $h$. The agent then produces a thought $t$ and a low-level action $a$ with type, position, and optional text input. For a multi-step task, each step $i$ has its own observation $o_i$, thought $t_i$, and action $a_i$, so the whole interaction between the environment and the GUI agent can be modeled as the following chain:
\[
(\mathcal{I}, (o_1, t_1, a_1), (o_2, t_2, a_2), \dots, (o_n, t_n, a_n))
\]

\subsection{RLVR \& GRPO Algorithm}

Reinforcement Learning with Verifiable Rewards (RLVR) algorithm allows LLMs to act as policy models $\pi(\theta)$, take states $s$, output actions $a$, and receive feedback on answer correctness from deterministic verifiers. Group Relative Policy Optimization (GRPO) algorithm, initially proposed in DeepSeekMath~\cite{shao2024deepseekmathpushinglimitsmathematical}, is one of the RLVR variants, which estimates advantages and updates the policy model while adhering to KL divergence constraints. For each response, the verifier assigns a reward $r_i$. We define a group of $N$ responses, with their rewards denoted as $\{r_1, r_2, ..., r_N\}$. The relative advantage $A_i$ of the $i$-th response is computed by:

\begin{equation}
    A_i = \frac{r_i-\text{mean}(\{r_1, r_2, ..., r_N\})}{\text{std}(\{r_1, r_2, ..., r_N\})}
\end{equation}

where mean and std denote the mean and standard deviation of the rewards. The loss of GRPO algorithm can be simplified as:

\begin{equation}
   \mathcal{L}_{GRPO}(\theta) = \mathbb{E}_i \left[ \frac{\pi_\theta(a|s)}{\pi_{\theta_{\text{old}}}(a|s)} A_i \right] - \alpha \cdot D_{KL}(\pi_{\theta_{\text{old}}} \parallel \pi_\theta) 
\end{equation}

Here, $s$ denotes current state, $\alpha$ is a balancing hyperparameter and $D_{KL}$ is the Kullback-Leibler divergence, which ensures the policy update within a reasonable range.

\section{Results on OmniACT}

\begin{table}[t]
\centering
\caption{Results on OmniACT. All experiments are conducted under the same zero-shot prompt. * denotes supervised fine-tuned on GUI-R1-3K~\cite{luo2025gui}. The best results are in bold.}
\label{tab:omniact}
\resizebox{\columnwidth}{!}{  
\begin{tabular}{lcccccc}
\toprule
\multirow{2}{*}{Models} & \multicolumn{3}{c}{OmniAct-Web} & \multicolumn{3}{c}{OmniAct-Desktop} \\
\cmidrule(lr){2-4} \cmidrule(lr){5-7}
 & Type & GR & SR & Type & GR & SR   \\ 
 \midrule
\textbf{Zero Shot}\\
\midrule
GPT-4o~\cite{gpt4oreport} & 79.33 & 42.79 & 34.06 & 79.97 & 63.25 & 50.67 \\ 
Qwen2.5-VL-3B~\cite{bai2025qwen25vltechnicalreport} & 50.63 & \colorbox{blue!10}{46.89} & 47.02 & 56.95 & \colorbox{blue!10}{47.97} & 46.89 \\  
Qwen2.5-VL-7B~\cite{bai2025qwen25vltechnicalreport} & 79.15 & 71.32 & 71.21 & 84.74 & 79.89 & 79.66 \\ 
\midrule
\textbf{Fine-Tuning}\\
\midrule
OS-Atlas-4B~\cite{wu2024atlas} & 46.74 & 49.24 & 22.99 & 63.30 & 42.55 & 26.94 \\ 
OS-Atlas-7B~\cite{wu2024atlas} & 85.63 & 69.35 & 59.15 & 90.24 & 62.87 & 56.73 \\ 
Qwen2.5-VL-3B*~\cite{bai2025qwen25vltechnicalreport} & 66.24 & 56.91 & 53.00 & 77.62 & 62.54 & 63.76 \\ 
Qwen2.5-VL-7B*~\cite{bai2025qwen25vltechnicalreport} & 81.62 & 73.45 & 73.39 & 86.23 & 80.17 & 79.80 \\ 
UI-R1-3B~\cite{lu2025uir1enhancingefficientaction} & 75.42 & 61.35 & 61.33 & 73.41 & 64.12 & 63.98 \\ 
GUI-R1-3B~\cite{luo2025gui} & 88.58 & 75.10 & 75.08 & 91.86 & 78.37 & 78.31 \\ 
\midrule
GUICrafter-3B (ours) & \textbf{89.22} & \textbf{77.21} & \textbf{77.37} & 91.36 & \textbf{82.88} & \textbf{82.88} \\
\quad -- only Stage1 & 88.63 & \colorbox{blue!10}{65.85 (18.96 $\uparrow$)} & 66.23 & \textbf{92.71} & \colorbox{blue!10}{78.81 (30.84 $\uparrow$)}  & 78.81 \\
\quad -- only Stage2 & 85.35 & 75.33 & 75.33 & 90.17 & 76.78 & 76.78 \\
\bottomrule
\end{tabular}
}
\end{table}

The results of GUICrafter on OmniACT are presented in Table \ref{tab:omniact}. Based on these results, we draw the following three conclusions:

\begin{itemize}
    \item After Stage 1, our model improves grounding accuracy by 18.96\% on the web domain and 30.84\% on the desktop domain compared with the base model Qwen2.5-VL-3B, highlighting the effectiveness of GUI pretraining.
    \item After both Stage 1 and Stage 2, our GUICrafter model outperforms other models of similar size on OmniACT. Since UI-TARS-2B has not released official results on OmniACT, we have not included it in Table \ref{tab:omniact}. On the web domain, our model exceeds the second-best GUI-R1 by 2.1\% in grounding accuracy, and on the desktop domain, it surpasses GUI-R1 by 4.5\%.
    \item After the curriculum learning process of Stage 1 and Stage 2, our model shows a 3.1\% improvement in grounding accuracy on the web domain and a 6.1\% improvement on the desktop domain compared to training with Stage 2 alone, which clearly confirms that GUI pretraining effectively strengthens the model.
\end{itemize}

\end{document}